\documentclass[11pt]{article}

\usepackage[final]{acl}

\usepackage{amsmath}
\usepackage{booktabs}
\usepackage{times}
\usepackage{latexsym}

\usepackage[T1]{fontenc}

\usepackage[utf8]{inputenc}

\usepackage{microtype}

\usepackage{inconsolata}

\usepackage{graphicx}

\title{Read Your Own Mind: Reasoning Helps Surface Self-Confidence Signals in LLMs}

\author{Jakub Podolak \\
  University of Amsterdam \\
  \small{\texttt{jakub.podolak.241 [at] gmail.com}} \\\And
  Rajeev Verma \\
  University of Amsterdam \\
}

\begin{document}
\maketitle
\begin{abstract}
We study the source of uncertainty in \texttt{DeepSeek R1-32B} by analyzing its self-reported verbal confidence on question answering (QA) tasks. In the default answer-then-confidence setting, the model is regularly over-confident, whereas semantic entropy---obtained by sampling many responses---remains reliable. We hypothesize that this is because of semantic entropy's larger test-time compute, which lets us explore the model's predictive distribution. We show that granting \texttt{DeepSeek} the budget to explore its distribution by forcing a long chain-of-thought before the final answer greatly improves its verbal score effectiveness, even on simple fact-retrieval questions that normally require no reasoning. Our analysis concludes that reliable uncertainty estimation requires explicit exploration of the generative space, and self-reported confidence is trustworthy only after such exploration.
\end{abstract}

\section{Introduction}

Generative language models (GLMs) like GPT, LLaMA, or Deepseek families have achieved great performance on diverse tasks \cite{dubey2024llama3herdmodels, deepseekai2025deepseekr1incentivizingreasoningcapability}, yet they are prone to failure modes such as ``hallucinations'' \cite{huang2023surveyhallucinationlargelanguage}. These inaccuracies can undermine trust and lead to poor decisions in LLM-assisted systems \cite{Huang_2024}. To mitigate this issue, quantification and the communication of model’s uncertainty in its outputs is seen as a potential to entrust these models with reliability.

\begin{figure}[t]
    \centering
    \includegraphics[width=\linewidth,
    clip,    
    trim=0 5pt 0 0]{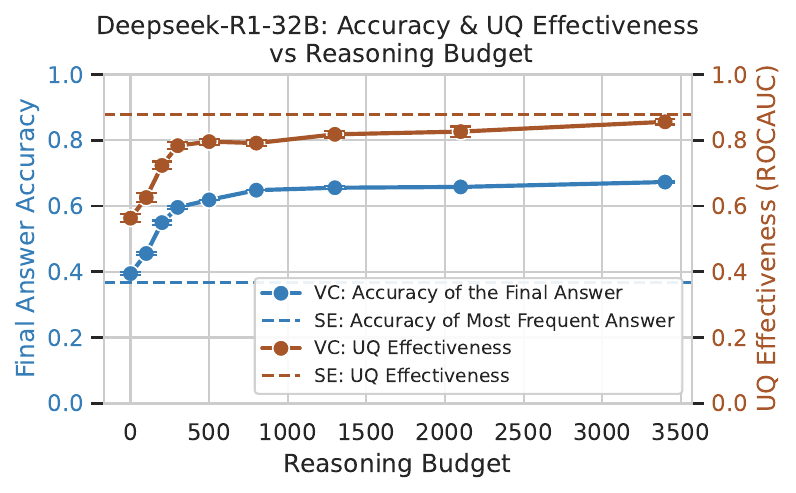}
    \caption{\textbf{DeepSeek R1-32B's Verbalized Confidence (VC) improves and matches Semantic Entropy's (SE) effectiveness, when longer reasoning is forced.}. Our work suggests that it is the test-time exploration of the model’s predictive space, not the particular uncertainty heuristic, that makes confidence estimates reliable.}
    \label{fig:figure-1-main-results}
\end{figure}

Numerous uncertainty-quantification (UQ) approaches have been proposed in this direction: from Monte-Carlo sampling based, such as Semantic Entropy (SE) \cite{farquhar2024detecting} to simpler Verbalized Confidence estimation (VC), which just asks the model directly to state its confidence \cite{xiong2024llmsexpressuncertaintyempirical}. 

While Verbalized Confidence estimation is easy to use, there is no scientific consensus on what it represents or what its source is, potentially making it unreliable to use in critical scenarios. Furthermore, prior work has shown that Verbalized Confidence is often pathological - for the same question a model might first return “Answer A (100 \% confidence)” and, in a second sample, “Answer B (95 \% confidence),” even though these probabilities cannot coexist \cite{xiong2024llmsexpressuncertaintyempirical}. On a more practical side, verbalized scores may be over-confident \cite{yang2024verbalizedconfidencescoresllms,pawitan2024confidencereasoninglargelanguage}, whereas Semantic Entropy remains comparatively well calibrated \cite{farquhar2024detecting}.

Semantic Entropy's effectiveness can be attributed to its test-time compute---allocating extra tokens at inference \cite{snell2024scalingllmtesttimecompute} to explore the predictive distribution. Test-time compute can also come in the form of an extended reasoning chain that precedes the final answer \cite{wei2023chainofthoughtpromptingelicitsreasoning,deepseekai2025deepseekr1incentivizingreasoningcapability}, and most recent works show that such reasoning can improve verbalized or token-level calibration as well \cite{zeng2025reasoningmodelsbetterverbalized, jurayj2025finalanswertesttimescaling}. These findings prompted us to pose a hypothesis: Is the model able to directly quantify and express verbally its uncertainty, or is the test-time token exploration necessary for the model to reliably summarize its confidence?

In this paper, we try to answer this question and better understand the source of VC by performing a set of experiments with \texttt{DeepSeek R1-32B} \cite{deepseekai2025deepseekr1incentivizingreasoningcapability} as a representative model. Our results show that without any chain-of-thought, \texttt{DeepSeek}’s verbalized scores carry little information about correctness. As we grant the model progressively larger reasoning budgets, its calibration improves and approaches the reliability of Semantic Entropy, even on simple fact-retrieval items. This trend suggests that meaningful uncertainty estimates emerge only after the model's predictive space has been explored, and that the final confidence percentage largely summarizes the diversity exposed in this process. We further enforce this hypothesis by using a separate reader model that, by just analyzing \texttt{DeepSeek}'s reasoning trace, matches the reliability of \texttt{DeepSeek}’s own Verbalized Confidence.

\section{Background and Related Work}

Generative language models frequently generate fluent but incorrect answers that can cause downstream harm \cite{band2024linguisticcalibrationlongformgenerations,Huang_2024}. When no external verifier is available, a model’s self-reported confidence is the only proxy for correctness, making reliable uncertainty estimates essential.


\paragraph{Calibration of LLM Confidence Scores.}
A confidence score is \emph{calibrated} if, for example, predictions tagged “80 \% confident” are correct roughly 80 \% of the time. Common approaches to obtain the confidence scores include token-level probabilities treated as a classification score \cite{dhuliawala-etal-2022-calibration}, semantic-level measures that evaluate agreement across multiple sampled completions \cite{farquhar2024detecting}, and explicitly verbalized percentages in a model’s output \cite{xiong2024llmsexpressuncertaintyempirical,tian2023justaskcalibrationstrategies}.  

\paragraph{Semantic Entropy vs. Verbalized Confidence}
A generative model, given a question \(Q\), defines a distribution over semantically distinct answers \(P(A \mid Q)\). The uncertainty of this distribution is naturally quantified by its Shannon entropy, and while computing it exactly is infeasible, we can approximate it by Monte-Carlo sampling and clustering semantically equivalent answers. This is exactly how the Semantic Entropy (SE) method \cite{farquhar2024detecting} works, leading to well-calibrated scores. The big downside of this method is that it requires sampling data on test-time (larger test-time-compute budget).

Another way to obtain a confidence score is simply to ask the model for one, for instance, ``I am 85 \% sure.''  This Verbalized Confidence (VC) is easy to collect and works with any black-box API \cite{xiong2024llmsexpressuncertaintyempirical,yang2024verbalizedconfidencescoresllms,ni2024largelanguagemodelshonest}.  
Yet, opposed to SE, it is unclear what the number represents: is the model sampling its own distribution, recalling similar training examples, or just guessing? To our best knowledge, no study has answered these questions, leaving the method too uncertain for safety-critical use.


The most recent works find that reasoning-tuned models that generate more tokens at the test time give better calibrated verbalized score \cite{hammoud2025answerreasoningtraceuncovers,wei2024measuringshortformfactualitylarge,xiong2024llmsexpressuncertaintyempirical,zhao-etal-2024-fact},
This hints that exploring test-time compute budgets' impact on Verbalized Confidence calibration might be crucial to understand its source, yet we are not familiar with any research work that tries to answer our questions specifically.

In this work, we systematically compare Verbalized Confidence and Semantic Entropy under matched test-time compute budgets, examine several task domains, and analyze the reasoning trace to see where the verbalized score comes from and why it lags behind Semantic Entropy.

\section{Methodology}

Our objective is to uncover where a model’s Verbalized Confidence comes from. We identify two competing views:

\textbf{Intrinsic latent variable:} the model can read out a hidden latent belief state and use it to express its uncertainty, and
\textbf{Self-sampling:} model does not have access to any reliable latent source of confidence, and reliable confidence emerges only after the model explicitly explores its own predictive space, as Semantic Entropy does by sampling many answers.

We test these views through a set of experiments that measure the behavior of VC when the model is forced to reason before answering, compare the effectiveness and accuracy to the SE, and analyze the uncertainty exposed in the reasoning traces. We describe our experimental setup in \autoref{appendix:experimental_setup}.

\section{Results}
Without any reasoning tokens, the score is barely better than random, and with enough exploration budget, VC can approach SE's effectiveness (\autoref{section:result_1}). Furthermore, an external reader can recover essentially the same uncertainty signal by inspecting the chain of thought alone (\autoref{section:result_3}), suggesting that the  Self-sampling hypothesis might be true.

\subsection{Extended Reasoning is Necessary for VC to Reach SE-Level Effectiveness}
\label{section:result_1}

\begin{figure*}[!t]
  \centering
  \includegraphics[
    width=0.9\textwidth,
    trim=0 20px 0 5px,
    clip
  ]{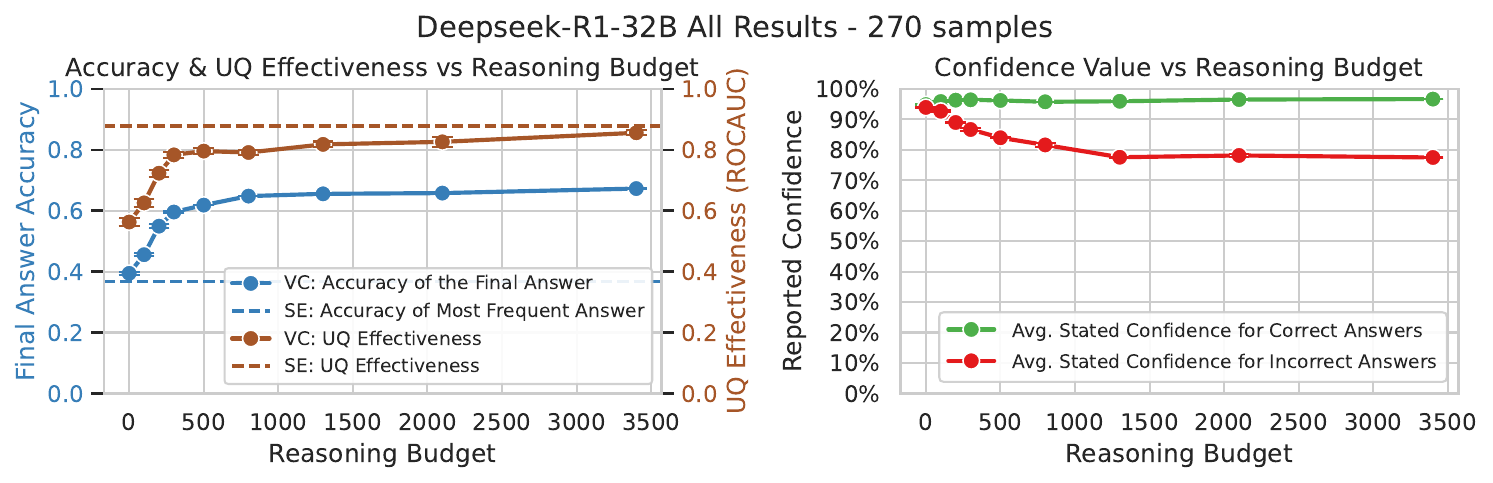}\\[1ex]
  \includegraphics[
    width=0.9\textwidth,
    trim=0 20px 0 0,
    clip
  ]{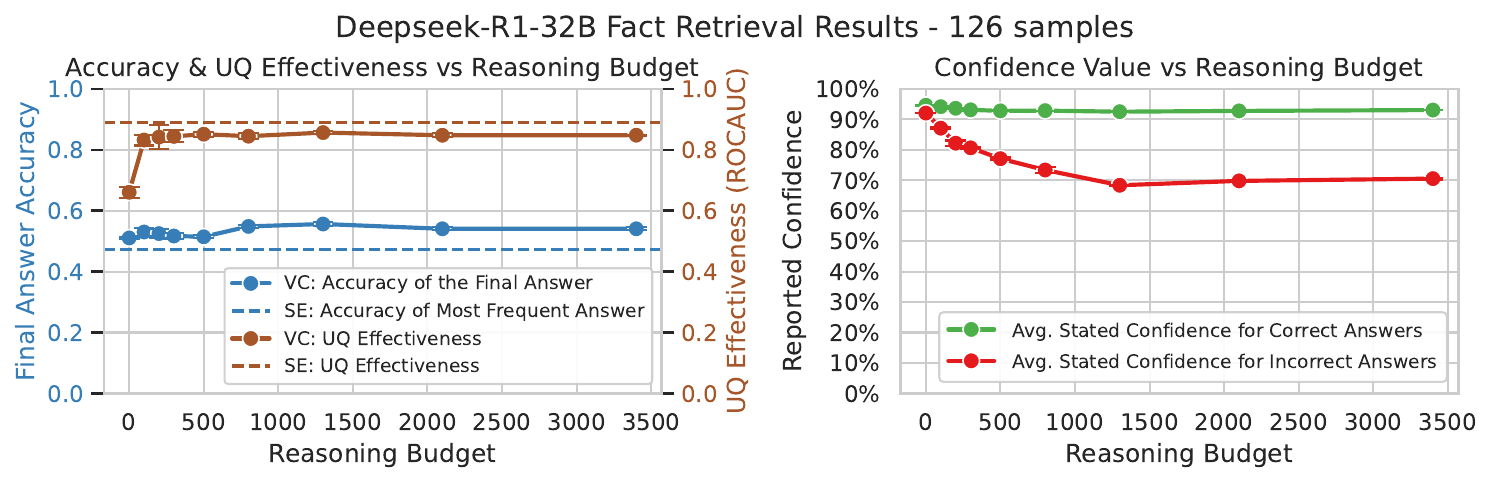}\\[1ex]
  \includegraphics[
    width=0.9\textwidth
  ]{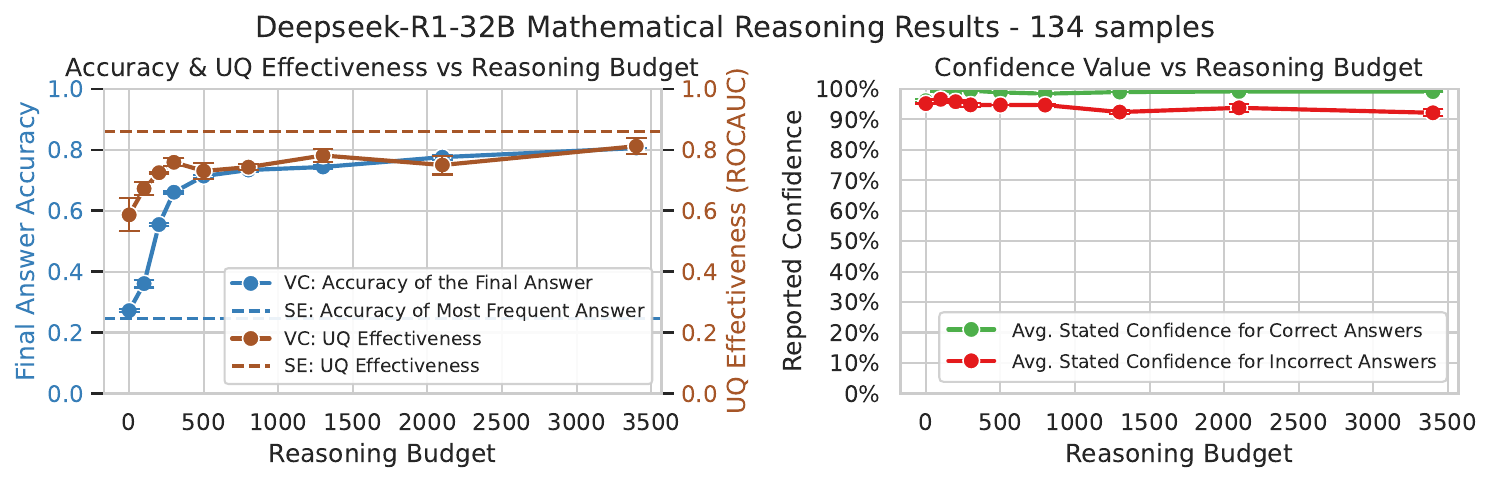}

  \caption{%
   \textbf{Effectiveness and Accuracy of Verbalized Confidence with Forced Reasoning vs Semantic Entropy.}
    (a) Full overview. 
    (b) Fact retrieval results. 
    (c) Mathematical reasoning results.
    Note: The remaining 10 samples not falling into the Fact Retrieval or Mathematical Reasoning categories are included in the Full overview but not presented as separate plots.
  }
  \label{fig:combined_three}
\end{figure*}

\autoref{fig:combined_three} shows final-answer accuracy, UQ effectiveness, and average stated confidence for correct and incorrect answers as a function of the reasoning budget, with Semantic Entropy shown for comparison. We see that granting just 100–500 reasoning tokens raises accuracy 41\% $\rightarrow$ 63\% and boosts verbalized-confidence ROC-AUC 0.56 $\rightarrow$ 0.80. 

For fact-retrieval questions (Fig.~\ref{fig:combined_three}b), answer accuracy does not improve with longer reasoning budgets, yet UQ effectiveness continues to improve with additional tokens. We can reach very long reasoning traces for fact-retrieval questions thanks to the employed forced reasoning technique \cite{muennighoff2025s1simpletesttimescaling} presented in \autoref{fig:setup}.

Verbalized Confidence is initially weaker than superior Semantic Entropy but reaches near parity at ~200 tokens for fact retrieval and ~3,500 tokens for mathematical items, while maintaining higher answer accuracy due to the reasoning process. For comparison, in our experiments, SE used 218 tokens per sample on average, meaning the two methods are very similar both in computational efficiency and UQ effectiveness for fact retrieval questions.

These results confirm that allocating test-time compute to reasoning is essential for reliable uncertainty estimates, and extended CoT effectively mitigates \texttt{DeepSeek}’s over-confidence without sacrificing performance. The sheer scale of the improvement in effectiveness: from near-random $0.56$ ROCAUC to $0.88$ suggests that there is no latent uncertainty information available for the model, and self-sampling is necessary to obtain a good uncertainty estimate.

\begin{figure}
    \centering
    \includegraphics[width=0.8\linewidth]{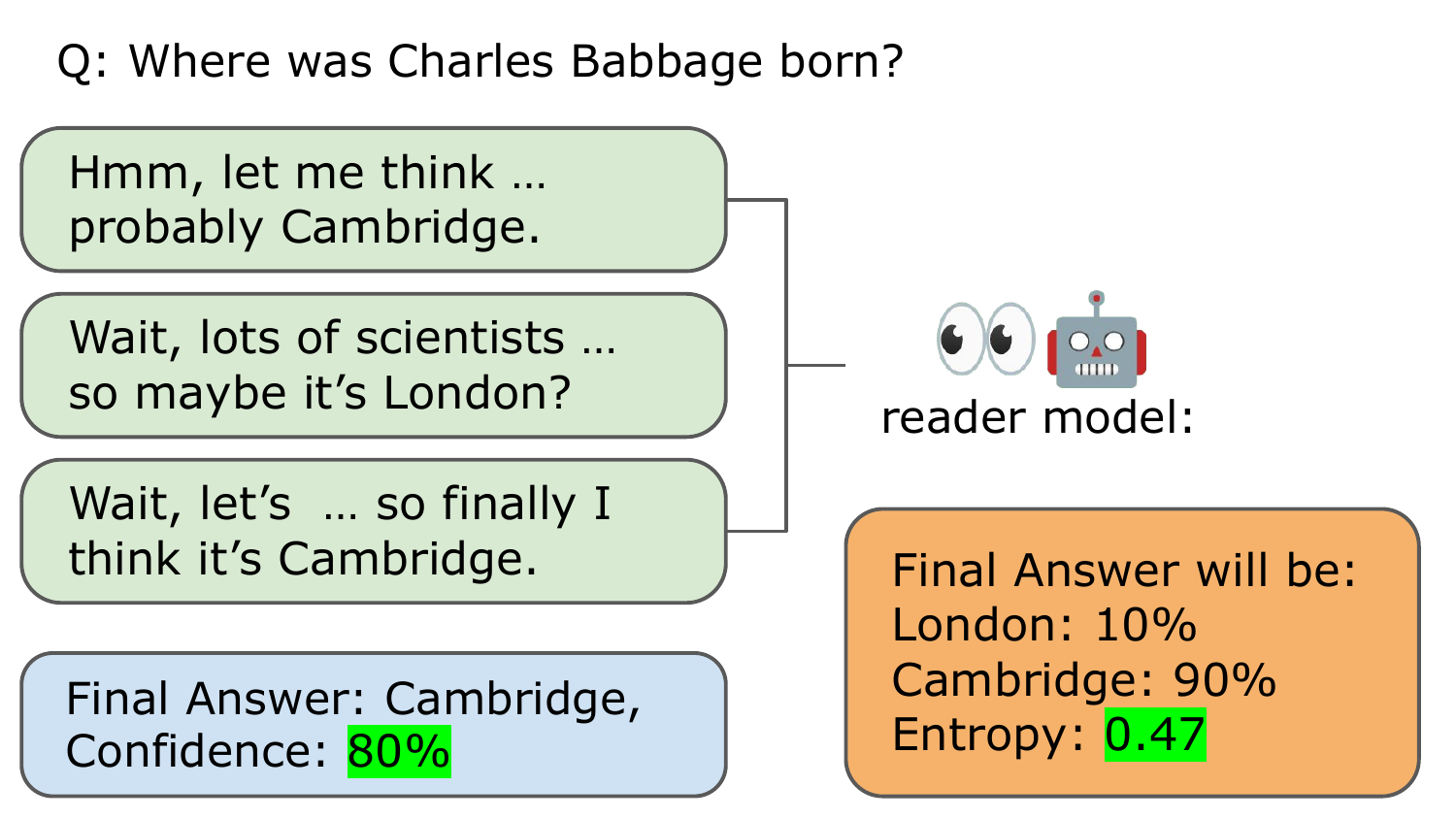}
    \includegraphics[width=0.8\linewidth]{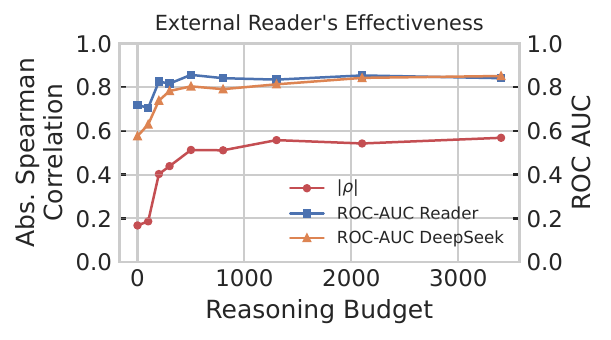}
    \caption{\textbf{Separate reader matches the reliability of DeepSeek’s own Verbalized Confidence by just looking at the reasoning trace.} With more reasoning tokens, the agreement between them (measured as absolute Spearman correlation) increases, and the effectiveness of both scores changes similarly.}
    \label{fig:reader-model-experiment}
\end{figure}

\subsection{External Reader Model Recovers VC Calibration from Reasoning Trace Alone}
\label{section:result_3}

If there is no hidden latent variable from which Verbalized Confidence is drawn, then the reasoning trace has to contain all the uncertainty information needed to explain \texttt{Deepseek}'s final score. We can verify it using an external reader model that, given only \texttt{DeepSeek}’s chain of thought, tries to predict its final answer and confidence.

\autoref{fig:reader-model-experiment} illustrates our experimental setup and results. As a reader model, we used OpenAI's GPT-4o-mini \cite{openai_gpt4o}, we provide more information about the setup in \autoref{appendix:prompts_and_inference}. We display (i) the absolute Spearman correlation $|\rho|$ between \texttt{DeepSeek}’s self-reported confidence and the reader entropy $H_{\text{reader}}$, and  (ii) the ROC-AUC of each score in detecting incorrect answers.

With no reasoning tokens exposed, the correlation between Reader's and \texttt{Deepseek}'s scores is low, however, with more reasoning tokens, the effectiveness of the reader goes up in tandem with \texttt{Deepseek}'s effectiveness, and the correlation between the two goes up.  At 3.4 k tokens, \texttt{DeepSeek} reaches ROC-AUC = 0.851 and the reader 0.841 with $|\rho|=0.57$, indicating that almost the entire confidence signal is now accessible in the trace.

These results support our claim that there is no directly accessible notion of uncertainty, and uncertainty information must be surfaced through test-time token sampling. When the model provides a Verbalized Score after the reasoning process, it most likely just reads its reasoning trace and summarizes the alternatives and uncertainty exposed in it.

\section{Discussion and Future Work}
We aimed to determine whether large language models can directly verbalize well-calibrated uncertainty or whether reliable confidence estimates only emerge after explicit exploration of their predictive space, via additional test-time compute such as parallel sampling (Semantic Entropy) or extended reasoning.

Our experiments suggest that  \textit{test-time compute, not the particular uncertainty heuristic, is the decisive factor for obtaining reliable confidence estimates in DeepSeek-R1-32B}.  
Left to produce only a short answer, the model remains over-confident because its belief state cannot be accessed directly.  
Granting the model additional tokens, either by sampling independent continuations (Semantic Entropy) or by forcing a longer chain of thought, allows it to externalize alternative hypotheses. This exposes a big issue with Verbalized Confidence - its appeal lies in the simplicity and how fast it is, yet it works well only after a significant reasoning computation is done.

While these results are encouraging, they may not be generalizable since we've tested only one model and used a very compact QA dataset. Furthermore, assuming it is true that test-time compute is the decisive factor for reliable UQ, it still might be the case that some methods of test-time compute may be more efficient in eliciting uncertainty than others. Future work could focus on making models reason more efficiently or explore their uncertainty in a more structured way. That could help Verbalized Confidence inherit the Semantic-Entropy-level of calibration with less computation needed.


\bibliography{custom}

\newpage
\appendix
\onecolumn

\section{Experimental Setup}
\label{appendix:experimental_setup}

\begin{figure*}
    \centering
  \includegraphics[
    width=1\textwidth,
    trim=0 103px 0 0,
    clip
  ]{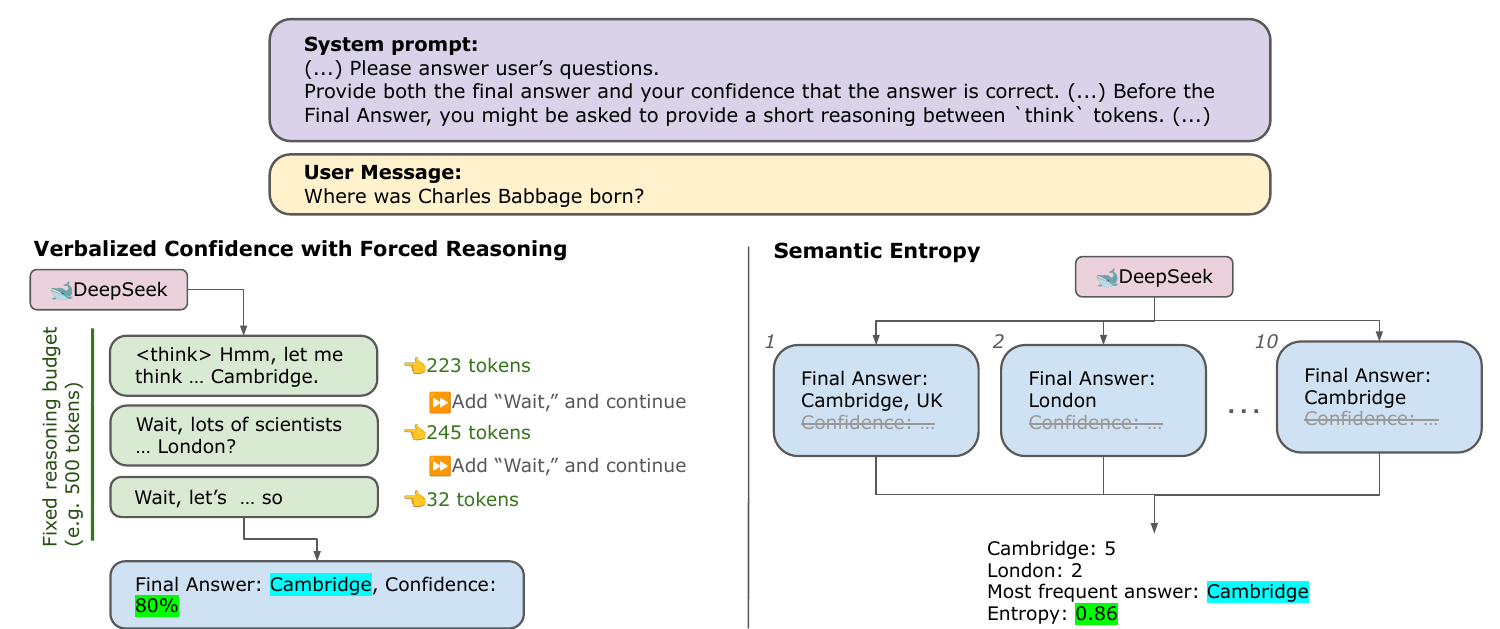}\\[1ex]
    \caption{\textbf{Two tested methods of obtaining Final Answer and Confidence} - Verbalized Confidence with Forced Reasoning (VC) works by prompting the model to reason for longer-until the fixed budget is exhausted - before stating the answer and confidence. Semantic Entropy (SE) obtains 10 independent answers that are later clustered semantically to identify the most frequent one, and to calculate the entropy in the answer distribution.}
    \label{fig:setup}
\end{figure*}

\paragraph{Data Sources.} Because long-trace experiments are computationally expensive, we built a small (270 samples) but diverse benchmark instead of using full datasets. We sampled questions from five popular, open-source sources: TriviaQA, MMLU, and SimpleQA for fact retrieval \cite{2017arXivtriviaqa,hendrycks2021measuringmassivemultitasklanguage,wei2024measuringshortformfactualitylarge}, plus GSM8K and AIME-2024 for mathematical reasoning \cite{cobbe2021trainingverifierssolvemath,hf_aime2024}.  
Our goal is open-ended QA in natural language, so we stripped away multiple-choice options in MMLU and any figure references in AIME-2024, manually discarding questions that could not stand alone after this edit, such as \textit{``Which of the following is true?''}.  
Every surviving example was then hand-labeled with its knowledge domain and the skills needed to answer it, such as ``Fact Retrieval'' or ``Mathematical Reasoning''. Full sampling details and the final label distribution appear in \autoref{appendix:data_creation_and_composition}.

\paragraph{Model and Prompts.} We chose \texttt{Deepseek‑R1‑32B}\footnote{\url{deepseek-ai/DeepSeek-R1-Distill-Qwen-32B}} \cite{deepseekai2025deepseekr1incentivizingreasoningcapability} following \citet{jurayj2025finalanswertesttimescaling} for its strong reasoning capabilities at a manageable model size. Furthermore, it is one of the most popular open-sourced reasoning-tuned models. All experiments were run on two NVIDIA A100 GPUs.

We provide an elaborate discussion on prompting and inference we adopted in \autoref{fig:setup}. Across all setups, we used a single system prompt that directs the model to (1) think step by step, and then (2) provide a final answer along with a Verbalized Confidence score. The full prompt text, as well as an interaction example, is available in \autoref{appendix:prompts_and_inference}. To regulate the length of the reasoning chain, we applied the budget‐based truncation method of  \citet{muennighoff2025s1simpletesttimescaling}: when the reasoning budget is exhausted (or set to zero), the chain terminates immediately. If the budget remains, the system appends “Wait, ” tokens, and asks to generate more tokens. For experiments with Verbalized Confidence, we lowered the decoding temperature to 0.1 to prevent the model from going off-topic in long reasoning. For parallel sampling in Semantic Entropy experiments, we set it to 1.0 to obtain more diverse responses and approximate the predictive distribution more efficiently.
 
\paragraph{UQ Methods.} Next, we describe how we obtain the estimates of verbalized score and semantic entropy: For Verbalized Confidence, we ask the model to provide the final answer and its confidence between 0\% and 100\% after (optional) forced reasoning (refer to \autoref{fig:setup} for visualization). 

For Semantic Entropy, we follow \citet{farquhar2024detecting}, and generate $n=10$ answers for each question with no reasoning chain. Afterwards, we use OpenAI's gpt-4o-mini \cite{openai_gpt4o} to cluster semantically equivalent generations. We select the majority cluster -- the cluster with the most members (i.e., the answer that appears most frequently once semantically equivalent responses are grouped) -- as the predicted answer and compute the Shannon entropy of the cluster-size distribution as the uncertainty score.

\paragraph{Evaluation and Metrics.} We report two main metrics: the accuracy of the final answer and the effectiveness of the UQ method measured as the area under ROC (ROC AUC) in the task of classifying the model's final answer correctness (hallucination classification). To calculate if the model's final answer is correct, we query OpenAI's GPT-4o-mini \cite{openai_gpt4o} model with the question if this proposed answer is equivalent to the ground truth answer in the dataset given a question.

We repeat experiments with verbalized score confidence across varying reasoning budgets 3 times and show mean and 95\% confidence intervals. For the rest of the experiments, we repeat them once unless stated otherwise.

\section{Data Creation and Composition}
\label{appendix:data_creation_and_composition}

Because our longest‐trace runs are expensive, we limited the benchmark to \textbf{270 open-ended questions} drawn from five well-known, permissively licensed QA datasets. We first sampled 310 items uniformly at random (seed 42) to balance fact-retrieval and mathematical-reasoning content while keeping the total below the $\approx$300-sample budget we could process. Items whose solutions required figures (AIME-2024), multiple-choice candidates (MMLU), or extra context passages (TriviaQA) were discarded after manual inspection, leaving the 270 used in all experiments (Table \ref{tab:before_after}). By doing so, we ensured that all the incorrect answers were caused by the model's mistakes, instead of missing context in the data.

\begin{figure}
    \centering
    \small
    \includegraphics[width=0.8\linewidth]{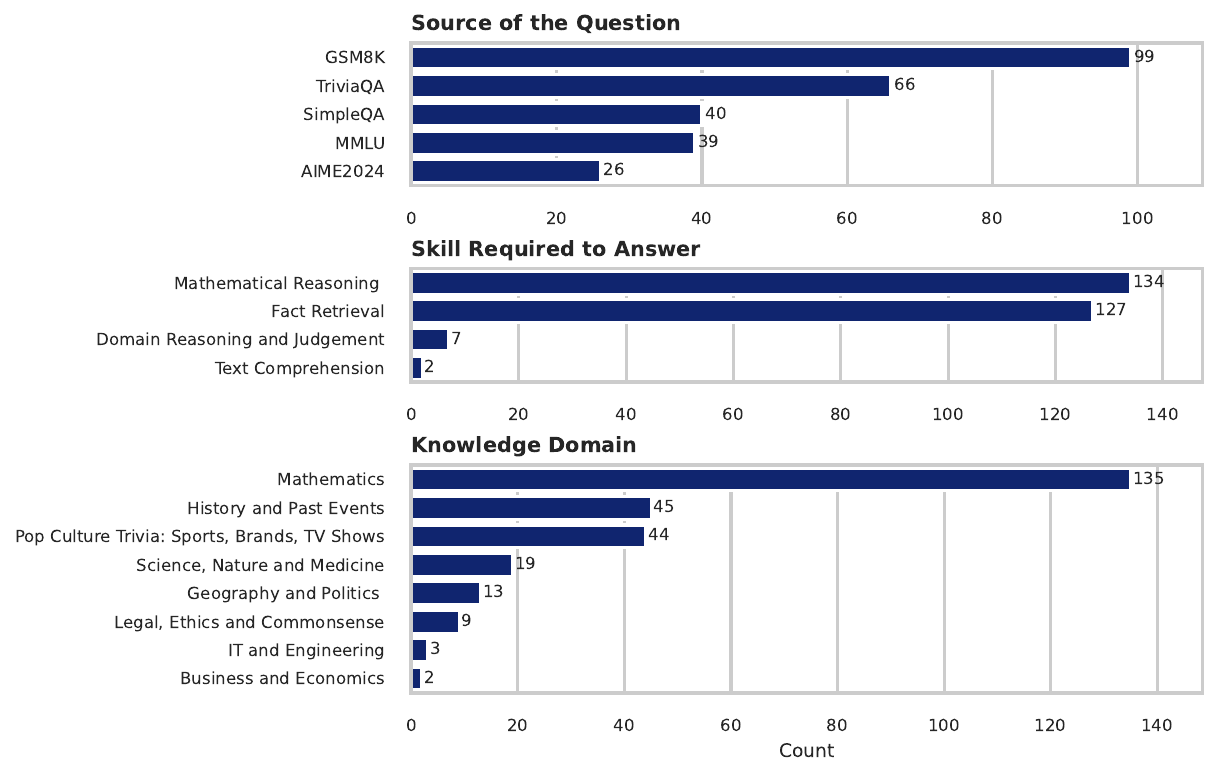}
    \caption{\textbf{Internal composition of our used data sample.}}
    \label{fig:internal_composition}
\end{figure}

Each example received two human labels - Knowledge Domain, and Skill Required. A large language model (OpenAI o3) proposed initial tags for 100 random questions. The first author then reviewed every instance, correcting tags where needed, and used these tags to manually label all 270 samples. You can find the specific tags and number of datapoints in \autoref{fig:internal_composition}. In the main paper, we break out results for the full dataset and for the two most common \emph{Skill Required} tags only; \emph{Knowledge Domain} splits are omitted because several categories are too small. Per-dataset results can be found in \autoref{appendix:results_across_samples}. Five representative questions and their tags are shown in \autoref{tab:sample_examples}.

\vspace{0.5em}
\begin{table}[h]
\centering\small
\setlength{\tabcolsep}{4pt}
\begin{tabular}{p{6cm} p{1.3cm} p{3.4cm} p{3.9cm}}
\toprule
\textbf{Example Question (truncated)} & \textbf{Dataset} & \textbf{Skill} & \textbf{Domain} \\
\midrule
In what year did Augustus De Morgan publish the article "Trochoidal Curve" in the Penny Cyclopaedia? & SimpleQA & Fact Retrieval & History and Past Events \\
\midrule
There exist real numbers $x$ and $y$, both greater than 1, such that $\log_x\left(y^x\right)=\log_y\left(x^{4y}\right)=10$. Find $xy$. & AIME2024 & Mathematical Reasoning  & Mathematics \\
\midrule
James runs 12 miles a day for 5 days a week.  If he runs 10 miles an hour how many hours does he run a week? & GSM8K & Mathematical Reasoning & Mathematics \\
\midrule
In Python 3, which of the following function removes all leading and trailing whitespace in string? & MMLU & Fact Retrieval & IT and Engineering \\
\midrule
Anaphylaxis is what sort of life-threatening illness? & TriviaQA & Fact Retrieval & Science, Nature and Medicine \\
\bottomrule
\end{tabular}
\caption{\textbf{Five representative items from the 270-question benchmark.}}
\label{tab:sample_examples}
\end{table}

\begin{table}[h]
\centering
\small
\begin{tabular}{lccc}
\toprule
\textbf{Dataset} & \textbf{Sampled} & \textbf{After Filtering} & \textbf{Removed} \\
\midrule
GSM8K      & 100 & 99  & 1  \\
TriviaQA   & 70  & 66  & 4  \\
SimpleQA   & 40  & 40  & 0  \\
MMLU       & 70  & 39  & 31 \\
AIME2024   & 30  & 26  & 4  \\
\midrule
\textbf{Total} & \textbf{310} & \textbf{270} & \textbf{40} \\
\bottomrule
\end{tabular}
\label{tab:before_after}
\caption{\textbf{Number of samples per dataset before and after manual filtering.}}
\end{table}

\section{Prompts and Inference}
\label{appendix:prompts_and_inference}

\begin{figure}
    \centering
    \small
    \includegraphics[width=0.9\linewidth]{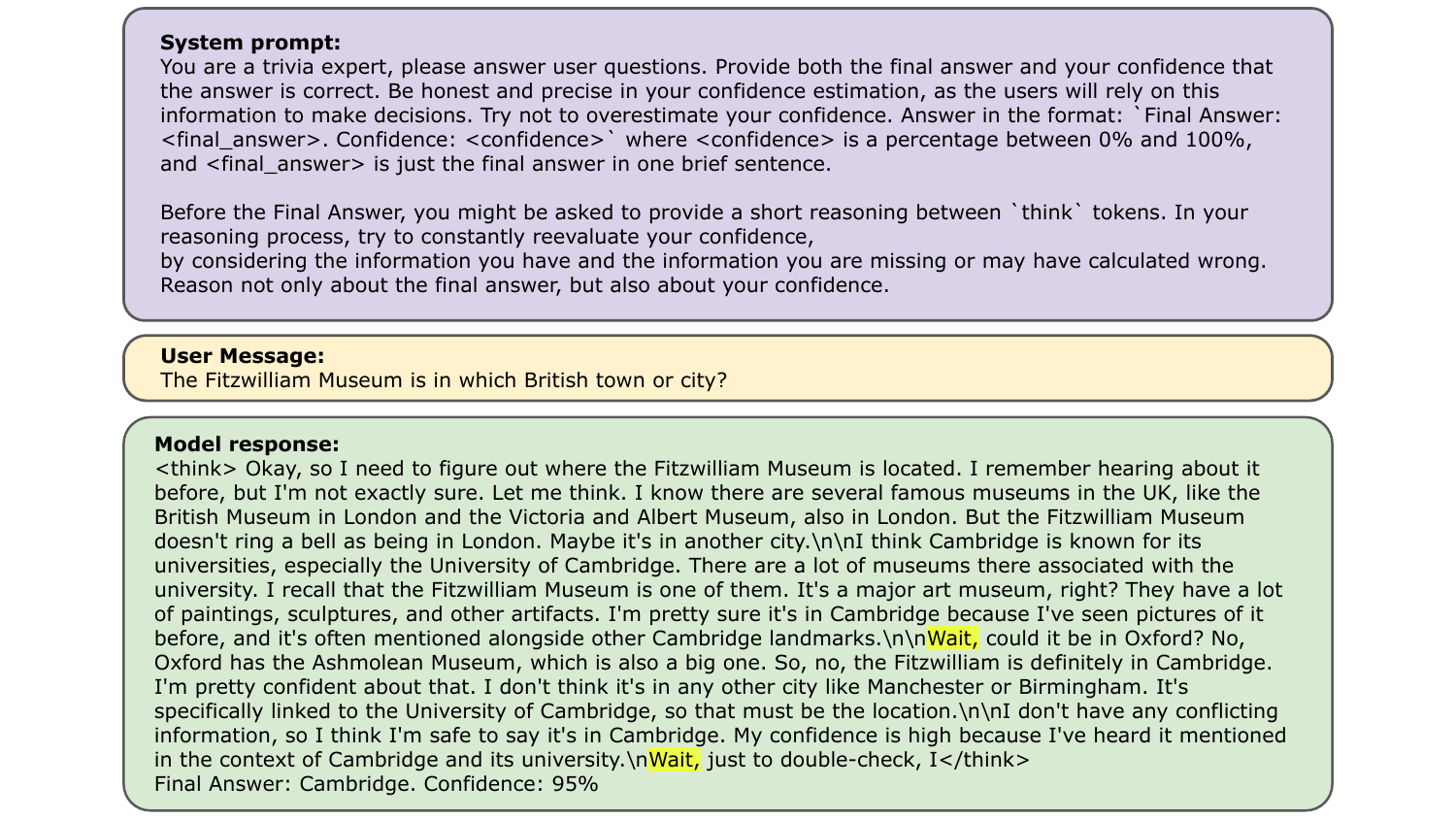}
    \caption{\textbf{Used system prompt and real interaction example}. The question comes from the TriviaQA dataset. Moments where the model was forced to reason for longer are highlighted.}
    \label{fig:real_interaction}
\end{figure}

\paragraph{Main VC vs SE runs.}
You can find the full system prompt used in the main VC vs SE experiments, as well as a real interaction example with Verbalized Confidence and budget forcing in \autoref{fig:real_interaction}.

\paragraph{Reader–model experiments.}
Our goal is to let an external model read DeepSeek’s reasoning trace and predict a probability distribution over possible DeepSeek's answers. After obtaining the distribution, we calculate Shannon entropy, which is used as a notion of uncertainty.
The procedure of obtaining a distribution over possible DeepSeek's answers has four steps:

\begin{enumerate}
\item \textbf{Candidate extraction.}  
For each question, we feed the entire $3.4\text{k}$-token reasoning chain to \texttt{gpt-4o-mini}, prompting it to list all candidate answers mentioned in the trace.
\item \textbf{Multiple-choice reformulation.}  
We label the distinct candidates with letters \texttt{A}, \texttt{B}, \texttt{C}\dots, we also add an option "None \ Unknown".
\item \textbf{One-token completion.}  
We prompt \texttt{gpt-4o-mini} again, asking which of the candidates is most likely to be returned by DeepSeek as a final answer. We ask the API for exactly one letter \texttt{A}, \texttt{B}, \texttt{C}\dots as a response. We keep the logits of this single-token prediction, apply \texttt{softmax}, and obtain a categorical distribution $p(A),p(B),\dots$.
\item \textbf{Reader entropy.}  
The Shannon entropy of that distribution, $H_{\text{reader}} = -\sum_i p(i)\log p(i)$, is the reader’s uncertainty score used in our analyses.
\end{enumerate}

You can find prompts that we used to extract the list of all candidates, as well as the final prediction, in figure \autoref{fig:reader_prompts}.

\begin{figure}[h]
  \centering
  \begin{minipage}{0.8\linewidth}
    \textcolor{blue}{
\textbf{Possible-candidates prompt}\\
}
``You are a helpful assistant. We asked a person to answer an open-ended question.
The person wrote a reasoning trace and then gave a final answer.
I want to know what all the possible options were that the person considered before giving the final answer.
This is for research on the answers people consider.
I want to use these possible answers to create a multiple-choice question.
Please give me all the possible answers the person considered, without duplicates, keeping the answers
distinct and suitable for a multiple-choice question.
Respond as a JSON list of strings, like this: FINAL LIST: ["answer1", "answer2", ...], which I can later use in a quiz.
Add the correct answer, the final answer, and Other / Unknown to the list, making sure there are no duplicates.
Reason about possible answers and how to remove duplicates, before giving the final list.
Include only answers mentioned in the reasoning trace plus the correct answer.
Do not add answers that are missing from the trace or the correct / unknown items.''

    \vspace{1em}
    \textcolor{teal}{
\textbf{Final-prediction prompt}\\
}

``You are a helpful assistant. We asked a person to answer a question; the person
started reasoning about the possible answers. Your role is to serve as an
autocompletion model that predicts the most likely final answer the person will give,
based on the current reasoning trace. You are not allowed to give your own answer—only the person’s most likely final answer.
Choose from the list of possible answers (A, B, C, …) and output exactly one letter,
with no whitespace or comments.''

    \vspace{0.5em}
\texttt{+ A) \{Option A\}}\\
\texttt{+ B) \{Option B\}}\\
\texttt{+ C) \{Option C\}}\\
\texttt{...}
  \end{minipage}
  \caption{\textbf{Reader-model prompts}. First prompt is used to extract candidate answers from the reasoning trace, the second one shows the follow-up prompt that asks the reader to predict DeepSeek's final choice. The last block of the second prompt is filled dynamically with the candidate list produced by the first one.}
  \label{fig:reader_prompts}
\end{figure}

\section{Results Across Datasets}
\label{appendix:results_across_samples}

\begin{figure*}[!t]
  \centering
  \includegraphics[
    width=0.8\textwidth,
    trim=0 20px 0 5px,
    clip
  ]{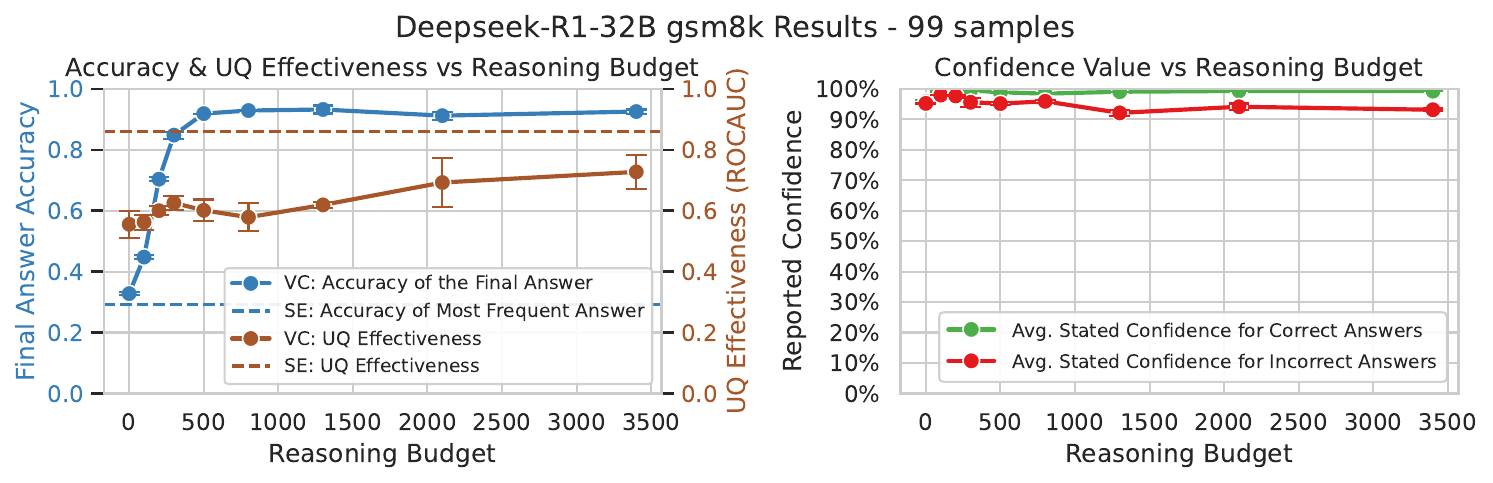}\\[1ex]
  \includegraphics[
    width=0.8\textwidth,
    trim=0 20px 0 0,
    clip
  ]{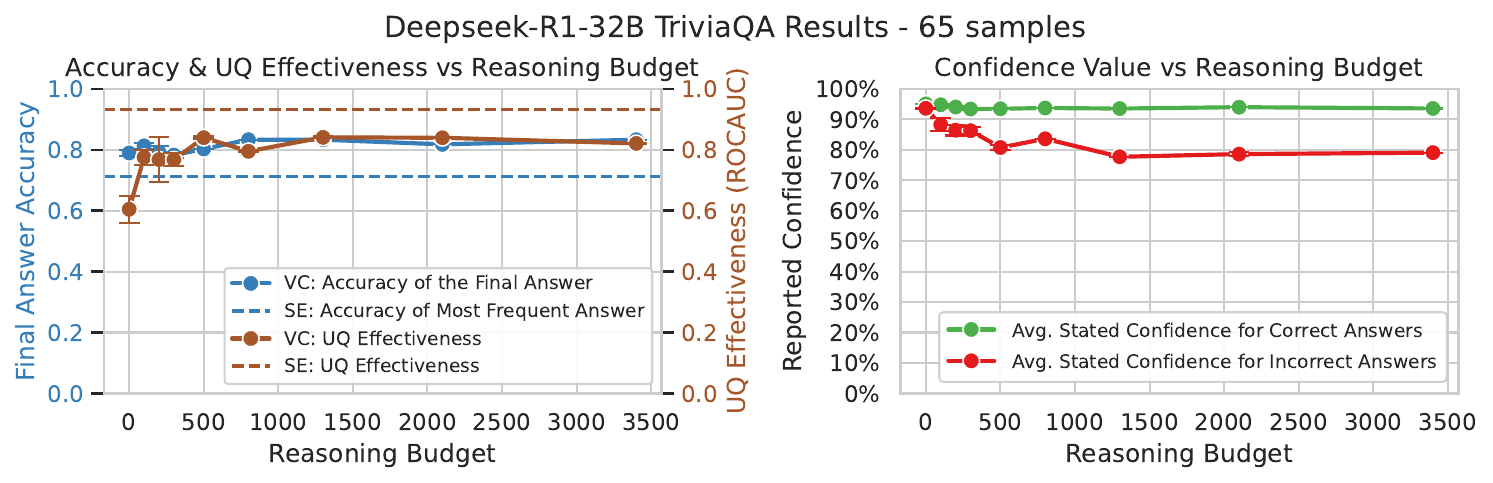}\\[1ex]
  \includegraphics[
    width=0.8\textwidth,
    trim=0 20px 0 0,
    clip
  ]{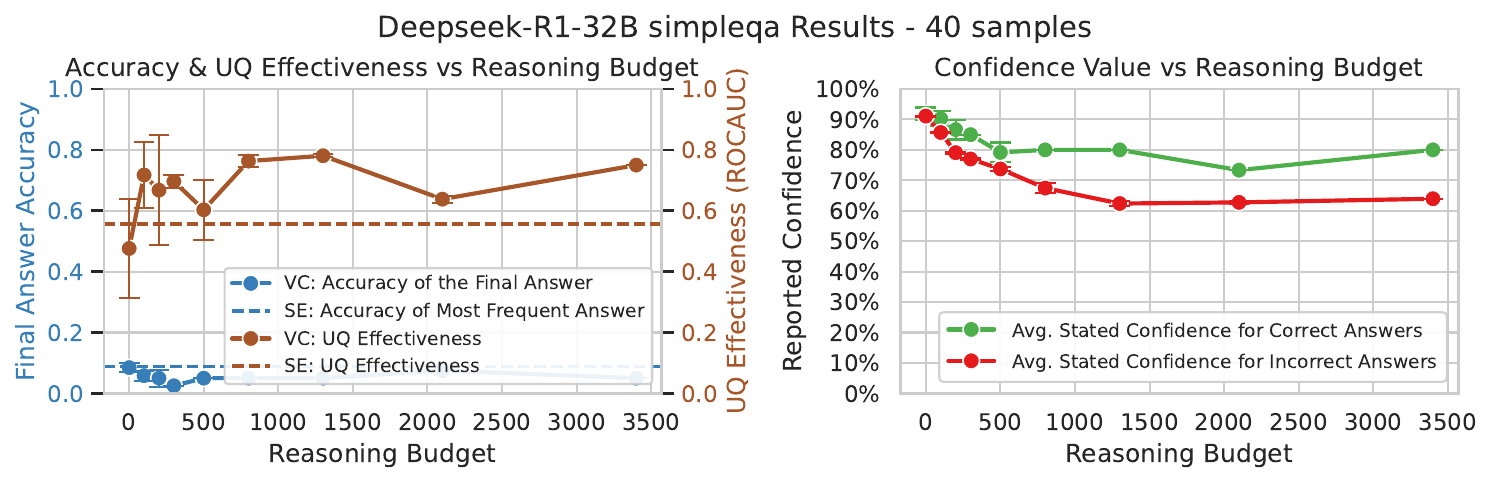}\\[1ex]
  \includegraphics[
    width=0.8\textwidth,
    trim=0 20px 0 0,
    clip
  ]{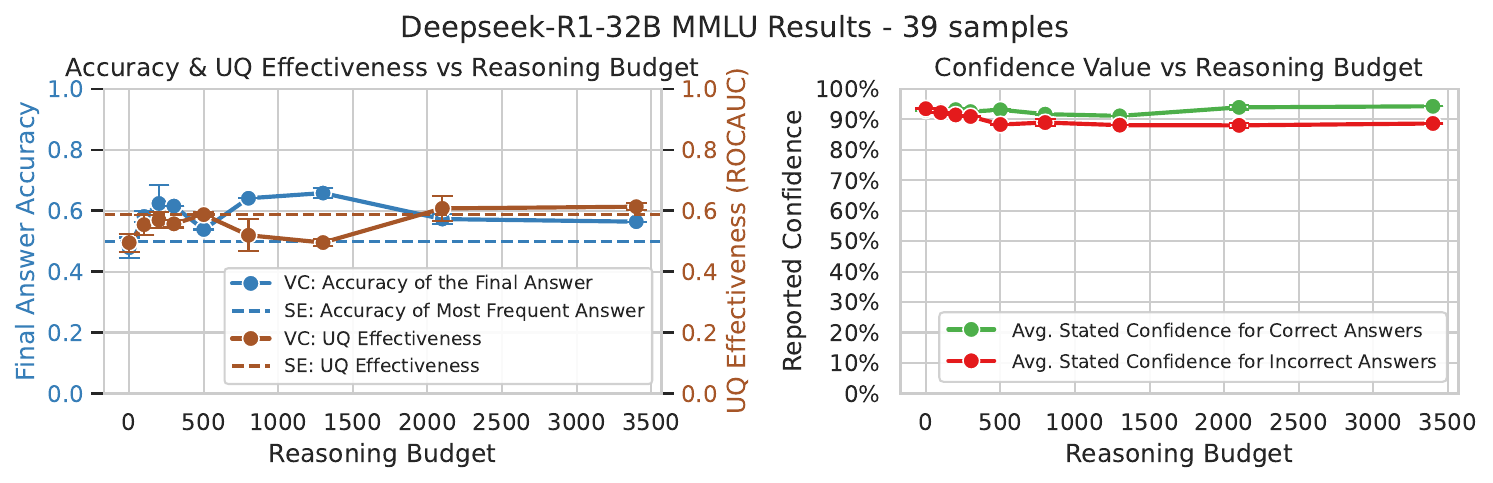}\\[1ex]
  \includegraphics[
    width=0.8\textwidth
  ]{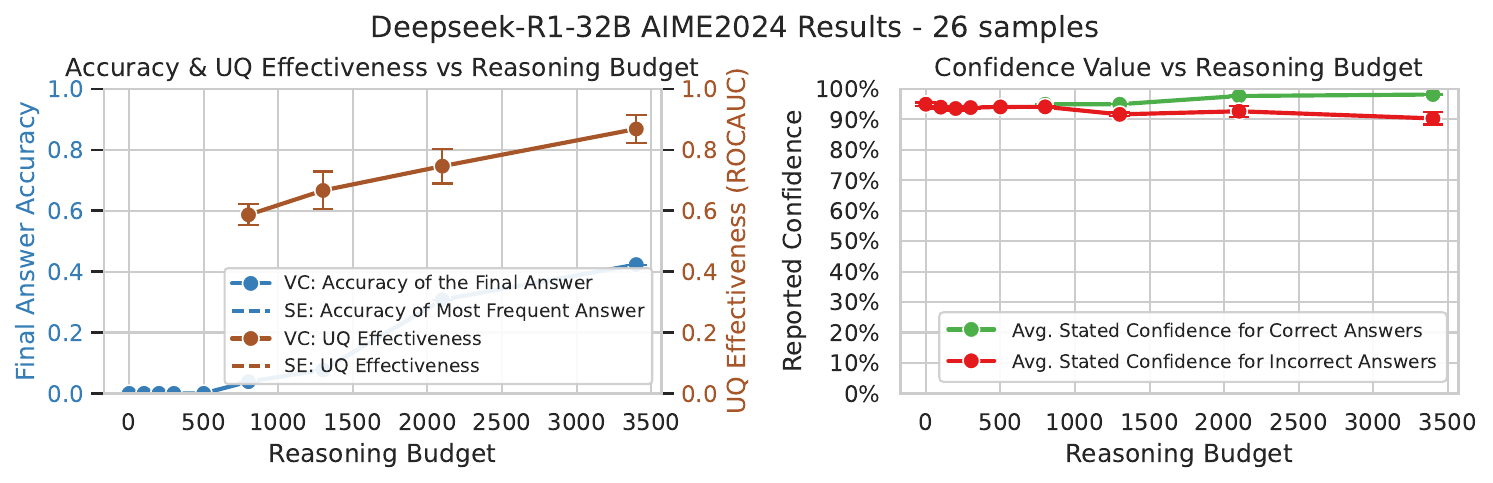}
  \vspace{-1.85pt}
  \caption{%
   \textbf{Effectiveness and Accuracy of Verbalized Confidence with Forced Reasoning vs Semantic Entropy.} Despite noise from limited samples, the right-hand plots show a consistent and increasingly pronounced divergence in reported confidence between correct and incorrect answers as the reasoning budget increases.
  }
  \label{fig:results_across_datasets}
\end{figure*}

We provide additional comparison of VC and SE with varying reasoning budgets across different sources of data in \autoref{fig:results_across_datasets}. While the trends are much noisier because of fewer samples, we see that for all the sources the difference in confidence between correct and incorrect answers increases with more reasoning tokens.

\end{document}